\documentclass[letterpaper, 10 pt, journal]{IEEEtran}

\usepackage{cite}
\usepackage{amsmath,amssymb,amsfonts}
\usepackage[caption=false,font=normalsize,labelfont=sf,textfont=sf]{subfig}
\usepackage{graphicx}
\usepackage{acronym}
\usepackage{textcomp}
\usepackage{xcolor}
\usepackage{booktabs}
\usepackage{multirow}
\usepackage{siunitx}
\usepackage{algorithm}
\usepackage[noend]{algpseudocode}
\algrenewcommand\algorithmicrequire{\textbf{Input:}}
\definecolor{Yellow}{rgb}{1,0.9,0.7}
\definecolor{Pink}{rgb}{1,0.85,0.85}
\definecolor{AntiqueWhite}{rgb}{0.9,0.9,0.9}

\acrodef{VLA}[VLA]{Vision-Language-Action}
\acrodef{MSE}[MSE]{Mean Squared Error}

\newcommand{\NOTE}[1]%
{
\noindent
\fboxsep=2mm\fcolorbox{black}{AntiqueWhite}{\parbox{0.90\columnwidth}
{\textbf{NOTE: } #1}
}
}

\usepackage{todonotes}
\usepackage[colorlinks=true,linkcolor=blue,citecolor=blue,urlcolor=blue]{hyperref}

\begin{document}
\bstctlcite{IEEEexample:BSTcontrol}
\title{Can Explicit Physical Feasibility Benefit VLA Learning? An Empirical Study\\
}
\author{Yubai Wei$^*$, Chen Wu, and Hashem Haghbayan
\thanks{$^*$Corresponding author.}
\thanks{The authors are with the Department of Computing, University of Turku, Finland (e-mail: yubwei@utu.fi; chen.c.wu@utu.fi; hashem.haghbayan@utu.fi).}
\thanks{This work has been submitted to the IEEE for 
  possible publication. Copyright may be transferred without notice, after which this version may no longer be accessible.}
}

\maketitle

\begin{abstract}
Vision-Language-Action (VLA) models map multimodal inputs directly to robot actions and are typically trained through large-scale imitation learning. While this paradigm has shown strong performance, prevailing VLA training procedures do not explicitly supervise hard physical constraints such as obstacle avoidance or kinematic feasibility. As a result, the geometric structure underlying physically feasible behavior must be inferred only implicitly from demonstrations. In this paper, we study whether introducing explicit feasibility supervision can provide effective structured guidance for VLA policies. We formulate a simple geometry-grounded feasibility objective and integrate it into the training stage of a diffusion-based VLA policy. To evaluate this idea systematically, we use obstacle-aware manipulation as a controlled probe of geometry-dependent physical feasibility. Empirical results show that augmenting VLA training with feasibility supervision improves both physical reliability and overall task performance, while also enhancing learning efficiency in the low-data regime. These findings indicate that explicit feasibility signals can effectively complement imitation-based VLA learning, highlighting their potential for developing more reliable VLA policies.

\end{abstract}


\section{Introduction}

Vision-Language-Action (VLA) models offer a promising path toward natural robot control: they map language instructions and visual observations directly to actions, bringing perception, reasoning, and control into a single policy. This paradigm has advanced rapidly through large-scale imitation learning on successful demonstrations, showing increasingly strong generalization across tasks, scenes, and embodiments~\cite{DBLP:conf/corl/ZitkovichYXXXXW23, DBLP:conf/rss/BrohanBCCDFGHHH23, DBLP:conf/rss/GhoshWPBMDHK0LT24, DBLP:conf/corl/KimPKXB0RFSVKBT24, DBLP:journals/corr/abs-2410-24164,DBLP:conf/iclr/LiuWLTCWX0025}. However, as VLA policies move closer to real deployment, task completion alone is not enough. In real-world manipulation, a robot must also satisfy hard physical constraints imposed by both the robot body and the environment, such as kinematic limits, self-collision avoidance, and obstacle avoidance~\cite{DBLP:books/cu/L2006, DBLP:journals/arcras/BrunkeGHYZPS22}. This raises a central question: \textit{Can VLA policies acquire physically feasible behavior, rather than merely actions that work in standard settings?}

Physical feasibility is fundamentally geometric. Whether a motion is feasible depends on the geometric relations among robot kinematics, body geometry, obstacle placement, and free space. Classical robotics treats this structure as first-class information through configuration-space reasoning, collision checking, and feasibility-aware search, which explicitly bias motion generation toward feasible solutions~\cite{DBLP:books/sp/90/Khatib90,Craig2005,DBLP:books/cu/L2006}. In contrast, prevailing end-to-end VLA pipelines typically do not include an analogous mechanism that explicitly represents and enforces such geometry-related feasibility structure during action generation~\cite{DBLP:journals/corr/abs-2512-11891,DBLP:conf/cvpr/WuZX0Y25}. 

\begin{figure}
    \centering
    \includegraphics[width=0.98\linewidth]{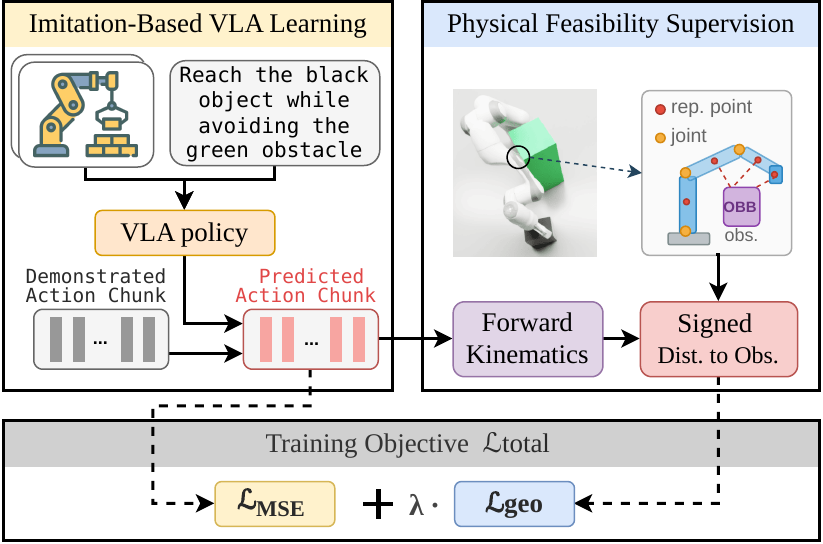}
    \caption{\textbf{Illustration on learning VLA with explicit physical feasibility supervision.} The policy is trained to match 
    demonstrated action chunks via $\mathcal{L}_{\mathrm{MSE}}$. We 
    additionally map predicted actions through forward kinematics and 
    compute signed distances to the obstacle, producing a feasibility 
    loss $\mathcal{L}_{\mathrm{geo}}$. The combined objective is 
    $\mathcal{L}_{\mathrm{total}}=\mathcal{L}_{\mathrm{MSE}}+\lambda 
    \mathcal{L}_{\mathrm{geo}}$. Obstacle geometry and kinematic 
    computations are used only during training; at inference the policy acts solely on RGB observations and language.} 
    \label{fig:method}
\end{figure}

As a result, in current VLA learning, the geometry underlying physical feasibility must largely be inferred from demonstrations. However, standard action-learning objectives supervise agreement with the demonstrated action sequence, rather than directly supervising geometric outcomes such as obstacle clearance, contact, or collision. A policy can therefore imitate successful behavior without being explicitly guided toward the geometric regularities that make it physically reliable.

In this paper, we empirically study explicit physical feasibility supervision in VLA learning and ask the following question: \textit{Can explicit physical feasibility supervision serve as an effective structured learning signal for VLA policies?} We introduce this supervision into the training process, and examine its effect using obstacle-aware manipulation as a controllable probe. Specifically, we consider close-obstacle reaching, where successful behavior requires both accurate target reaching and sufficient robot-obstacle clearance throughout execution. Within this setting, we augment a diffusion-based VLA policy with a simple geometry-grounded feasibility objective during training, which is illustrated in Fig.~\ref{fig:method}. Our simulated experiments show that, in this controlled setting, explicit feasibility supervision can improve physical reliability and task accuracy simultaneously, enhance learning efficiency under limited data, and be most effective when applied with appropriate strength. These findings suggest that physical feasibility can serve as a structured learning signal for VLA policies, pointing toward a broader opportunity to incorporate physical priors into the data-driven VLA learning paradigm.

Our main contributions are as follows:
\begin{itemize}
\item We formulate physical feasibility supervision as a learning problem for VLA policies, and design a controlled experimental setup using close-obstacle reaching as a probe where geometry-dependent feasibility is directly measurable.
\item We instantiate this idea with a minimal differentiable feasibility objective, designed to isolate the effect of physical supervision from imitation learning.
\item We provide empirical evidence that explicit feasibility supervision serves as a structured and complementary learning signal for training VLA policies.
\end{itemize}

\section{Related Work}
\subsection{Advances and Emerging Challenges in VLA}

Recent breakthroughs in VLAs have demonstrated their potential in high-level task execution and reasoning~\cite{DBLP:conf/corl/ZawalskiCPMFL24,DBLP:conf/cvpr/0003XKISGX24,DBLP:conf/cvpr/ZhaoLKFZWLMHFHL25}, with advances spanning cross-task generalization, complex instruction-following, cross-embodiment adaptation~\cite{DBLP:journals/corr/abs-2505-06111}, and long-horizon planning~\cite{DBLP:journals/corr/abs-2507-04447}. 

Despite these advances, bridging the gap from high-level capabilities to real-world deployment presents a fundamental challenge: the strict physical constraints of embodied interaction~\cite{DBLP:conf/corl/IchterBCFHHHIIJ22,DBLP:conf/corl/HuangWZL0023}. Beyond performing well on a given task, deployed policies must satisfy hard physical feasibility to avoid concrete physical hazards, such as unintended collisions~\cite{DBLP:journals/arcras/BrunkeGHYZPS22,DBLP:journals/jair/ZhaoLHL25,kim2026modularsafetyguardrailsnecessary}. Recognizing this critical gap, recent VLA works have begun to explicitly address physical safety and reliability. For instance, SafeVLA~\cite{DBLP:journals/corr/abs-2503-03480} addresses VLA safety through constrained reinforcement learning, CoFreeVLA~\cite{DBLP:journals/corr/abs-2601-21712} introduces explicit risk estimation for collision-aware execution and refinement, and AEGIS/VLSA~\cite{DBLP:journals/corr/abs-2512-11891} equips VLA systems with a plug-and-play safety layer for constraint-aware deployment. This trend highlights physical feasibility and safety as a key focus for developing future VLA systems.

\subsection{Physical Feasibility as Learning Signals}
Physical feasibility has long been fundamental to robot control and planning, where kinematic constraints and collision avoidance are essential for real-world deployment~\cite{DBLP:books/sp/90/Khatib90,Craig2005,DBLP:books/cu/L2006}. More recently, physical feasibility has also been explored as a learning signal in learned robot policies. This line of work includes neural motion-planning methods such as M$\pi$Nets~\cite{DBLP:conf/corl/FishmanMEPBF22} and its transformer-based successor Avoid Everything~\cite{DBLP:conf/corl/FishmanWBYSBF24}, as well as recent generative visuomotor policies, including diffusion-based approaches such as RK-Diffuser~\cite{DBLP:conf/cvpr/0006PHJ24} and KStar Diffuser~\cite{DBLP:conf/cvpr/0001LD0LHGWN25}. These works suggest that collision-aware or kinematics-aware signals can improve the reliability and quality of the learned policy. Together, these studies indicate that physical feasibility can be a useful source of supervision in robot policy learning.

However, the integration of such signals within the prevailing VLA paradigm remains limited~\cite{DBLP:journals/access/KawaharazukaOYPZ25}. Existing VLA systems have been driven primarily by large-scale data-driven learning, focusing on mapping visual observations and language instructions to action predictions, while physical feasibility is typically not explicitly encoded in the learning objective~\cite{DBLP:conf/corl/ZitkovichYXXXXW23, DBLP:conf/rss/BrohanBCCDFGHHH23, DBLP:conf/rss/GhoshWPBMDHK0LT24, DBLP:conf/corl/KimPKXB0RFSVKBT24, DBLP:journals/corr/abs-2410-24164,DBLP:conf/iclr/LiuWLTCWX0025}. Motivated by this gap, our work studies the effect of injecting explicit physical feasibility supervision into VLA learning, offering a complementary, learning-centered perspective on how such supervision shapes policy learning beyond imitation alone.

\section{Learning VLA with Feasibility Supervision}
In this section, we describe the formulation used to study 
physical feasibility supervision in VLA policy learning. 
We first define the task setting, then review the diffusion-based 
VLA formulation, and finally introduce the geometry-grounded 
feasibility objective.

\subsection{Task Definition}
\label{sec:task}
To study physical feasibility, we require a setting in which physical constraints are tightly coupled to policy performance. We therefore consider a close-obstacle reaching task, a representative near-boundary setting in robot motion planning. In this task, a single-arm policy receives visual observations and a language instruction, and must reach a specified target while operating in the presence of a nearby obstacle. Accordingly, task success requires both reaching the target within a tolerance and maintaining sufficient geometric clearance from the obstacle throughout execution. In this setting, physical feasibility is naturally characterized by the robot–obstacle clearance, making the task a controlled testbed for feasibility-aware VLA learning.

\subsection{Diffusion-based VLA policy}
\label{sec:diffusionpolicy}
In this study, we follow a standard diffusion-based formulation for VLA policy learning~\cite{DBLP:conf/rss/ChiFDXCBS23,DBLP:conf/iclr/LiuWLTCWX0025}. Given a language instruction \(\ell\) and visual observations \(\mathbf{o}_t\), the policy models the conditional distribution:
\begin{equation}
    p(\mathbf{a}_{t:t+T_a-1} \mid \ell, \mathbf{o}_t).
\end{equation}

In robotic manipulation, \(\mathbf{a}_{t:t+T_a-1}\) denotes a sequence of future actions, i.e., an action chunk:
\begin{equation}
\mathbf{a}_{t:t+T_a-1} := (\mathbf{a}_t, \ldots, \mathbf{a}_{t+T_a-1}),
\end{equation}
where \(T_a\) denotes the chunk size~\cite{DBLP:conf/rss/ZhaoKLF23, DBLP:conf/iclr/LiuWLTCWX0025}. For notation simplicity, \(\mathbf{a}_t^0\) denotes the clean action chunk starting at time \(t\), and \(\mathbf{a}_t^k\) its noisy version at diffusion step \(k\), where \(k \in \{1,\dots,K\}\) and \(K\) is the total number of denoising steps.

At inference time, action generation starts from Gaussian noise, \(\mathbf{a}_t^K \sim \mathcal{N}(\mathbf{0}, \mathbf{I})\), and is iteratively denoised into a clean action chunk \(\mathbf{a}_t^0\) through \(K\) denoising steps conditioned on the multimodal inputs. During training, the standard paradigm follows denoising-based imitation learning, which directly learns the demonstrated action chunk from noisy observations. Given an expert action chunk \(\mathbf{a}_t^0\), a forward diffusion process perturbs it with Gaussian noise \(\epsilon_k \sim \mathcal{N}(\mathbf{0}, \mathbf{I})\) at a randomly sampled diffusion step \(k \sim \mathcal{U}(\{1,\dots,K\})\):
\begin{equation}
\mathbf{a}_t^k
=
\sqrt{\bar{\alpha}_k}\,\mathbf{a}_t^0
+
\sqrt{1-\bar{\alpha}_k}\,\epsilon_k ,
\end{equation}

where \(\bar{\alpha}_k\) is the cumulative noise schedule coefficient at diffusion step \(k\). A denoising network \(F_\theta\) is then trained to recover the clean expert action chunk directly from the noisy input:
\begin{equation}
\hat{\mathbf{a}}_t^0 = F_\theta(\mathbf{a}_t^k, \ell, \mathbf{o}_t, k).
\end{equation}
The standard learning objective is the \ac{MSE} between the predicted and expert action chunks:
\begin{equation}
\mathcal{L}_{\text{MSE}}
=
\mathbb{E}_{\mathbf{a}_t^0,\epsilon_k,k,\ell,\mathbf{o}_t}
\left[
\left\|
\mathbf{a}_t^0 - F_\theta(\mathbf{a}_t^k,\ell,\mathbf{o}_t,k)
\right\|_2^2
\right].
\end{equation}

\begin{figure*}[t]
    \centering
    \includegraphics[width=\textwidth]
    {picture/generation_data_illustration_final_v2.pdf}
    \caption{Synthetic data generation pipeline for obstacle-avoidance episodes. The pipeline consists of object generation under multi-view visual validation and counterfactual trajectory generation, which constructs an obstacle-avoidance trajectory by re-planning from an obstacle-free reference trajectory. Only the obstacle-present trajectory is retained in the final dataset.}
    \label{fig:data pipeline}
    \vspace{-8pt}

\end{figure*}

\subsection{Physical Feasibility Supervision}
\label{sec:feasibilityloss}

While the standard \ac{MSE} objective learns action imitation from demonstrations, it does not explicitly enforce physical feasibility. To address this, we introduce an auxiliary geometric supervision signal grounded in two sources of information available at training time: robot kinematics and obstacle geometry. In our collision-avoidance setting, distance to the obstacle naturally serves as an explicit physical feasibility signal. Such geometric distance objectives have been widely used for collision evaluation and trajectory optimization in classical robotics~\cite{DBLP:conf/icra/RatliffZBS09,DBLP:journals/ijrr/SchulmanDHLABPPGA14}. Furthermore, recent work has shown that differentiable distance representations can provide smooth, useful gradients for motion generation and optimization, and can be integrated into control and learning-based pipelines~\cite{DBLP:conf/iros/LiuZTJ0C22, DBLP:conf/rss/OrtizC0SNZM22,DBLP:conf/icra/LiZRC24, DBLP:conf/rss/LiCRC24}. Based on this established formulation, we adopt a simplified differentiable distance-based supervision built on forward kinematics and signed distance to the obstacle, allowing geometric signals to directly guide policy learning through backpropagation.

\paragraph{Forward Kinematics Mapping}
 We first convert the predicted action chunk into future joint states over the prediction horizon. Forward kinematics then maps each future joint state to the corresponding robot link poses in the workspace:
\begin{equation}
\mathbf{T}_{t+\tau}^{(l)} = f_{\mathrm{FK}}^{(l)}(\mathbf{q}_{t+\tau}), \qquad \tau=0,\ldots,T_a-1,
\end{equation}
where \(\mathbf{q}_{t+\tau}\) is the predicted joint state at future step \(\tau\) within the denoised action chunk $\hat{\mathbf{a}}_t^0$, and \(\mathbf{T}_{t+\tau}^{(l)}\) denotes the pose of link \(l\). This deterministic mapping converts the policy outputs in joint space into geometric information of the rigid robot body.

\paragraph{Signed Distance to Obstacle}
Based on the link poses, we evaluate the geometric relationship between the robot and the obstacle. In practice, instead of densely sampling the full robot surface, we use the origins of a small set of selected links, denoted $\mathcal{S}$, as representative robot points. Each obstacle is modeled as an oriented bounding box (OBB), denoted by $\mathcal{B}_t$, which is used for signed-distance computation during training. For each representative point, we compute its analytic signed distance to the OBB surface. To account for body thickness, we further subtract a link-specific radius:

\begin{align}
\mathbf{p}_{t+\tau}^{(l)} &=\mathbf{T}_{t+\tau,1:3,4}^{(l)}, \\
d_{t+\tau}^{(l)} &= \mathrm{SDF}_{\mathrm{OBB}}\!\left(\mathbf{p}_{t+\tau}^{(l)};\mathcal{B}_t\right)-r^{(l)}.  
\end{align}

Here, \(\mathbf{p}_{t+\tau}^{(l)}\) is the representative point of link \(l\), \(r^{(l)}\) is its associated radius, and \(d_{t+\tau}^{(l)}\) denotes the resulting surface clearance. Positive, zero, and negative values indicate clearance, contact, and penetration, respectively.

\begin{algorithm}[t]
\caption{Training with Physical Feasibility Supervision}
\footnotesize
\label{alg:training}
\begin{algorithmic}[1]
\Require Dataset $\mathcal{D}=\{(\ell,\mathbf{o}_t,\mathbf{a}_t^0,\mathcal{B}_t)\}$, denoising network $F_\theta$, representative link set $\mathcal{S}$, link radii $\{r^{(l)}\}_{l\in\mathcal{S}}$, safety margin $\delta$, loss weight $\lambda$, action horizon $T_a$, total diffusion steps $K$
\For{each training iteration}
    \State Sample \label{alg:imitation_begin}
    $(\ell,\mathbf{o}_t,\mathbf{a}_t^0,\mathcal{B}_t)\sim\mathcal{D}$
    \State Sample $k\sim\mathcal{U}(\{1,\dots,K\})$, $\epsilon_k\sim\mathcal{N}(\mathbf{0},\mathbf{I})$
    \State $\mathbf{a}_t^k \gets \sqrt{\bar{\alpha}_k}\,\mathbf{a}_t^0 + \sqrt{1-\bar{\alpha}_k}\,\epsilon_k$ \Comment{Forward diffusion}
    \State $\hat{\mathbf{a}}_t^0 \gets F_\theta(\mathbf{a}_t^k,\ell,\mathbf{o}_t,k)$ \Comment{Denoising prediction}
    \State $\mathcal{L}_{\mathrm{MSE}} \gets \|\mathbf{a}_t^0-\hat{\mathbf{a}}_t^0\|_2^2$ \Comment{Imitation loss} \label{alg:imitation_end}

    \For{$\tau = 0,\dots,T_a-1$} \Comment{Forward kinematics mapping} \label{alg:kinematics_begin}
    \State Extract $\mathbf{q}_{t+\tau}$ from $\hat{\mathbf{a}}_t^0$ 
        \For{each $l\in\mathcal{S}$}
            \State $\mathbf{T}_{t+\tau}^{(l)} \gets f_{\mathrm{FK}}^{(l)}(\mathbf{q}_{t+\tau})$
            \State $\mathbf{p}_{t+\tau}^{(l)} \gets \mathbf{T}_{t+\tau,1:3,4}^{(l)}$
            \State $d_{t+\tau}^{(l)} \gets \mathrm{SDF}_{\mathrm{OBB}}(\mathbf{p}_{t+\tau}^{(l)};\mathcal{B}_t)-r^{(l)}$ \Comment{Signed distance}
        \EndFor
    \EndFor

    \State $\mathcal{V} \gets \{(\tau,l)\mid d_{t+\tau}^{(l)}<\delta\}$ \Comment{Active violations}
    \If{$|\mathcal{V}|>0$}
        \State $\mathcal{L}_{\mathrm{geo}} \gets \frac{1}{|\mathcal{V}|}\sum_{(\tau,l)\in\mathcal{V}}\left[\max(0,\delta-d_{t+\tau}^{(l)})\right]^2$
    \Else
        \State $\mathcal{L}_{\mathrm{geo}} \gets 0$
    \EndIf
    \Comment{Geometric feasibility loss} \label{alg:kinematics_end}

    \State $\mathcal{L} \gets \mathcal{L}_{\mathrm{MSE}} + \lambda\,\mathcal{L}_{\mathrm{geo}}$ \label{alg:update_begin}
    \State Update $\theta$ by descending $\nabla_\theta \mathcal{L}$ \label{alg:update_end}
\EndFor
\end{algorithmic}
\end{algorithm}

\paragraph{Geometric Feasibility Loss}

We apply a squared hinge loss to the representative link points across the predicted action chunk. For each future step \(\tau\) and link \(l\), the loss is activated when the corresponding surface clearance \(d_{t+\tau}^{(l)}\) falls below a safety margin \(\delta\). Here, \(\delta\) defines the range of geometric supervision: a larger \(\delta\) causes more near-obstacle points to be penalized, while a smaller \(\delta\) focuses the supervision on more critical near-collision cases. To prevent the loss from being diluted by safe links and future steps, we average the loss only over active violations. Let $\mathcal{V}=\{(\tau,l)\mid d_{t+\tau}^{(l)}<\delta\}$ denote the set of active violations. The geometric feasibility loss is defined as:
\begin{equation}
\mathcal{L}_{\mathrm{geo}}
=
\operatorname{Avg}_{\,d_{t+\tau}^{(l)}<\delta}
\left[\max\!\left(0,\delta-d_{t+\tau}^{(l)}\right)\right]^2.
\end{equation}
The overall training objective is
\begin{equation}
\mathcal{L}_{\mathrm{total}}
=
\mathcal{L}_{\mathrm{MSE}}+\lambda\,\mathcal{L}_{\mathrm{geo}},
\end{equation}
where \(\lambda\) controls the relative strength of physical feasibility supervision compared to \ac{MSE} term. 

Notably, the obstacle geometry and forward kinematics used to compute $\mathcal{L}_{\mathrm{geo}}$ are required \emph{only} during training. At inference time, the policy receives only RGB observations and the language instruction; no explicit obstacle geometry, signed-distance queries, or forward-kinematics computations are involved in action generation. The feasibility objective therefore acts as a training-time inductive bias that encourages the policy to internalize geometry-aware behavior from visual inputs alone, without introducing any additional perception or calibration requirements at deployment. 
The overall training procedure of Algorithm~\ref{alg:training} is  as follows:
\begin{itemize}
    \item \textit{Imitation learning (Lines~\ref{alg:imitation_begin}--\ref{alg:imitation_end}):} a diffusion step is sampled and Gaussian noise is injected into the expert action to obtain a noisy action, which is then denoised by the network, and compute the imitation loss $\mathcal{L}_{\text{MSE}}$.
    \item \textit{Geometric feasibility (Lines~\ref{alg:kinematics_begin}--\ref{alg:kinematics_end}):} the predicted action sequence is mapped to joint configurations through forward kinematics, link-wise signed distances to obstacles are evaluated, and a hinge-based penalty is applied to violations to obtain geometric feasibility loss $\mathcal{L}_{\mathrm{geo}}$.
    \item \textit{Update parameters (Lines~\ref{alg:update_begin}--\ref{alg:update_end}):} the imitation and feasibility objectives are combined, and the network parameters are updated.
\end{itemize}

\section{Obstacle-avoidance Data Generation}
\label{sec:pipeline}

In Sec.~\ref{sec:task}, we define our task as a close-obstacle reaching problem. In this section, we present the simulation pipeline used to construct the dataset in NVIDIA Isaac Sim~\cite{nvidia_isaac_sim}. Specifically, we implement a single-arm scene with Franka Arm, where targets and obstacles are randomly spawned for each episode to collect data. For each episode, we generate a target and an obstacle, assign a corresponding language instruction, plan executable trajectories using MoveIt with OMPL~\cite{sucan2012ompl} as the motion planner, and record synchronized RGB observations from three camera views: an overview camera, a left wrist-mounted camera, and a fixed right-side camera, together with the action trajectory. The recorded action trajectories and obstacle parameters also enable the subsequent derivation of geometric feasibility signals when needed. As illustrated in Fig.~\ref{fig:data pipeline}, the pipeline mainly consists of two components, object generation and counterfactual trajectory generation.

\subsection{Object Generation}
In each episode, we randomly generate a target and an obstacle with randomized poses and colors sampled from a predefined color set. All objects are modeled as cubes, so that their geometry can be consistently represented as OBBs for subsequent feasibility signal computation.

To ensure that the visual observations provide sufficient spatial information for reasoning, we enforce both visibility and separability constraints during scene generation. Using the known intrinsics and extrinsics of the camera, we project the OBB of each object onto the image plane for subsequent conditional sampling. A scene is accepted only if (i) the OBBs of the target and obstacle do not intersect in 3D space, and (ii) in at least two camera views, both objects are fully within the image frame and remain clearly separable. Scenes that violate any of these conditions are rejected and resampled.

\subsection{Counterfactual Trajectory Generation}
When constructing close-obstacle trajectories, we require the obstacle to lie close to an obstacle-free reference trajectory so that it meaningfully interferes with the original motion, while still allowing a collision-free re-plan to the same goal with only a local deviation from the reference trajectory.

Accordingly, we use a counterfactual trajectory generation procedure to place the obstacle and construct the corresponding obstacle-avoidance trajectory. For each episode, we first plan a reference trajectory to the target in an obstacle-free scene, denoted as \(\pi^{-}\). We then generate an obstacle and place it near the swept path of \(\pi^{-}\) by resampling its pose until the minimum robot-obstacle clearance along \(\pi^{-}\) falls below a small threshold \(\epsilon\). This ensures that the obstacle would collide with, or come very close to, the reference \(\pi^{-}\). With the obstacle fixed, we re-plan from the same start state to the same goal using the same planner settings, producing a collision-free trajectory \(\pi^{+}\) in the obstacle-present scene. The reference trajectory \(\pi^{-}\) is used only for obstacle construction and is not kept in the final dataset.

\section{Experiments}
In this section, we investigate whether explicit physical feasibility supervision can serve as a useful structured learning signal for VLA policies. We organize our empirical study around three research questions. \emph{RQ1:} How does explicit feasibility supervision affect policy behavior? \emph{RQ2:} Does explicit feasibility supervision improve learning efficiency under limited data? \emph{RQ3:} How does supervision strength influence the effectiveness of explicit feasibility supervision? 

We address these questions through simulation experiments in NVIDIA Isaac Sim simulation environment using a single Franka Arm robot~\cite{haddadin2022franka} with the obstacle-aware manipulation setting introduced above, which serves as a controlled probe of geometry-dependent physical feasibility.

\subsection{Experimental Setup}
Here, we describe our experimental setup, covering the policy backbone and compared methods, training data collection, obstacle perturbation protocol for evaluation, metrics, and implementation details.

\paragraph{Backbone and Compared Methods}
We adopt RDT-1B~\cite{DBLP:conf/iclr/LiuWLTCWX0025}, the largest open-source diffusion-based foundation model for robotic manipulation (1.2B parameters), as the policy backbone. We compare two fine-tuning objectives: \begin{itemize}
    \item \emph{MSE} (baseline): standard denoising-based imitation learning with the \ac{MSE} objective $\mathcal{L}_{\mathrm{MSE}}$ only.
    \item \emph{MSE+Feasibility} (ours): the combined objective $\mathcal{L}_{\mathrm{total}} = \mathcal{L}_{\mathrm{MSE}} + \lambda\,\mathcal{L}_{\mathrm{geo}}$, which augments imitation with explicit physical feasibility supervision.
\end{itemize}
All other configurations are identical to isolate the effect of the feasibility signal.

\paragraph{Training Data}

We collect data through the pipeline in Sec.~\ref{sec:pipeline} with clearance threshold $\epsilon = 0.10$\,m. Each episode records three-view RGB observations, the expert obstacle-avoidance trajectory planned by OMPL~\cite{sucan2012ompl}, and a language instruction generated by Gemini-Robotics-ER-1.5-Preview (e.g., ``Reach for the black object on your left, avoiding the green obstacle''). Table~\ref{tab:dataset_stats} summarizes the full 120-episode dataset; we use subsets of 40, 80, and 120 episodes to study data efficiency.

\begin{table}[t]
\centering
\caption{Statistics of the training dataset. Distances in this table are reported in centimeters (cm) for readability.}
\label{tab:dataset_stats}
\begin{tabular}{ll}
\toprule
Statistic & Value \\
\midrule
Total episodes & 120 \\
RGB views per episode & 3 \\
Sampling frequency & 15\,Hz \\
Steps per episode & 80 \\
\midrule
$d_{\min}$ (mean $\pm$ std) & $6.57 \pm 3.11$\,cm \\
$d_{\mathrm{tgt}}$ (mean $\pm$ std) & $8.14 \pm 6.60$\,cm \\
$\Pr(d_{\min} < 2\,\text{cm})$ & 6.50\% \\
$\Pr(d_{\min} < 5\,\text{cm})$ & 31.71\% \\
$\Pr(d_{\mathrm{tgt}} < 10\,\text{cm})$ & 89.43\% \\
$\Pr(d_{\mathrm{tgt}} < 15\,\text{cm})$ & 91.06\% \\
\bottomrule
\end{tabular}
\end{table}

\begin{figure}[t]
    \centering
    \includegraphics[width=0.98\columnwidth]{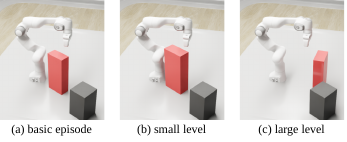}
    \caption{Illustration of obstacle perturbations used at evaluation. The gray cube denotes the target, which remains fixed, while the red cube denotes the obstacle, whose placement is perturbed. Starting from a basic episode, we construct small and large perturbation levels by changing the obstacle position while keeping the start state, target, and language instruction unchanged.}
    \label{fig:perturbation}
\end{figure}

\begin{figure*}[t]
    \centering
    \includegraphics[width=0.95\textwidth]{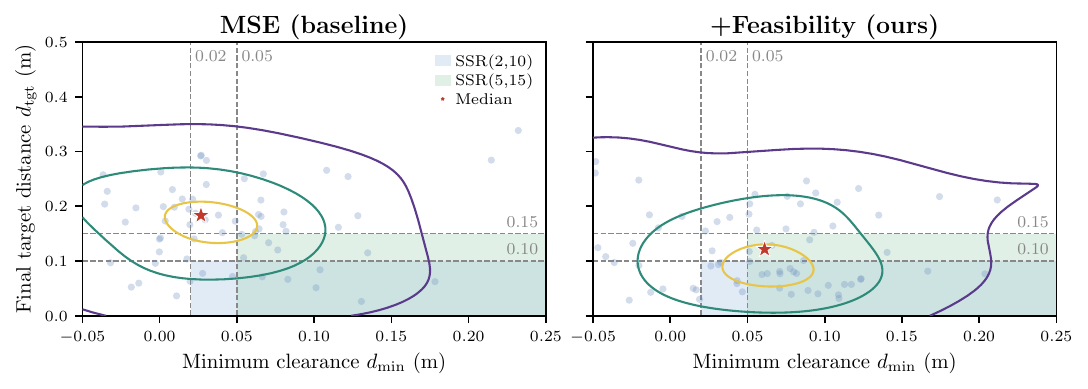}
    \caption{Joint distribution of $d_{\min}$ and $d_{\mathrm{tgt}}$ 
    under large perturbations (40 training episodes). 
    Shaded regions indicate the two SSR criteria. 
    Contour lines show 10\%, 50\%, and 90\% KDE density levels; 
    dashed lines mark the SSR thresholds (values annotated). 
    With feasibility supervision, the distribution shifts toward 
    higher clearance and lower target error simultaneously.}
    \label{fig:joint_dist}
\end{figure*} 
\begin{figure}[t]
    \centering
    \includegraphics[width=\columnwidth, trim=0.2cm 0.2cm 0.4cm 0cm, clip]{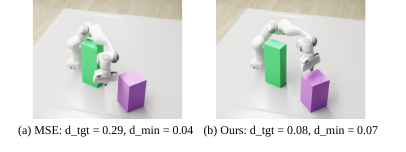}
    \caption{Qualitative comparison under a large obstacle perturbation. 
    The \ac{MSE} baseline nearly contacts the obstacle with poor reaching 
    accuracy, while the policy with feasibility supervision maintains 
    larger clearance and reaches closer to the target.}
    \label{fig:case}
\end{figure}

\paragraph{Evaluation under Obstacle Perturbations}
Since the task requires obstacle-aware actions, we perturb the obstacle configuration at test time while keeping the start state, target, and language instruction fixed. This allows us to probe whether the policy adjusts its actions in response to changed obstacle placements rather than simply replaying the demonstrated trajectory. We consider two perturbation levels (Fig.~\ref{fig:perturbation}):
\begin{itemize}
    \item \emph{Small perturbations} apply local changes around the original obstacle placement, including random translational shifts of 0-0.10\,m in the $xy$-plane and slight size variations. These are used to evaluate robustness under in-distribution obstacle configurations.
    \item \emph{Large perturbations} relocate the obstacle to a substantially different yet still trajectory-interfering position, producing much larger visual changes in the observations. These are used to evaluate local generalization under out-of-distribution obstacle configurations.
\end{itemize}

\paragraph{Evaluation Metrics}
We assess \emph{safety} via the minimum signed robot-obstacle clearance
$d_{\min}$ (negative values indicate collision) and \emph{task accuracy}
via the final hand-to-target distance $d_{\mathrm{tgt}}$. The
\emph{Safe Success Rate}
$\mathrm{SSR}(\alpha,\beta)=\Pr(d_{\min}>\alpha \;\wedge\; d_{\mathrm{tgt}}<\beta)$
jointly captures both.

We report two complementary variants:
$\mathrm{SSR}(0.05,0.15)$, which emphasizes \emph{safe approach}
by requiring a more comfortable safety margin while relaxing precision;
and $\mathrm{SSR}(0.02,0.10)$, which emphasizes \emph{precise reaching}
by tightening the reaching tolerance while allowing near-contact
obstacle avoidance. As both quantities are continuous, we additionally report
$\Pr(d_{\min}<0.02)$ and $\Pr(d_{\min}<0.05)$ for safety, and
$\Pr(d_{\mathrm{tgt}}<0.10)$ and $\Pr(d_{\mathrm{tgt}}<0.15)$ for
accuracy, providing a fuller characterization of the behavior
distribution. Unless otherwise noted, all distances in the following analysis are reported in meters (m).

\paragraph{Implementation Details}

We perform full-parameter fine-tuning of RDT-1B on two NVIDIA A100 GPUs with a learning rate of $1 \times 10^{-4}$ for 2\,000 epochs across all configurations. Training wall-clock time is approximately 8, 15, and 23 hours for 40, 80, and 120 episodes, respectively. Both the training and prediction action chunk size are $T_a = 64$ at 15\,Hz. At execution time, the policy predicts and fully executes one chunk at a time for 3 chunks per episode (192 steps in total). 
 
For the feasibility objective, we use $\delta = 0.10$\,m and $\lambda = 1.0$ as default hyper-parameters; the ablation over different $(\delta, \lambda)$ combinations is presented in Sec.~\ref{sec:rq3}. Forward kinematics and OBB signed-distance computations are implemented as differentiable operations so that $\mathcal{L}_{\mathrm{geo}}$ provides gradients to the policy network through backpropagation.
 
At evaluation time, we adopt different rollout protocols for the two
perturbation levels, reflecting their different roles. For
small perturbations, which represent in-distribution
variations, we sample 5 perturbations per base episode and perform 1
rollout per perturbation. For large perturbations, which
represent out-of-distribution obstacle relocations, we sample 2
perturbations per base episode and perform 5 rollouts each, averaged
per perturbation. We report mean performance across perturbations,
with 95\% CI half-widths on SSR estimated via episode-level clustered bootstrap.

\subsection{RQ1: Behavior Change under Obstacle Perturbations}
\label{sec:rq1}

\begin{table}[t]
\centering
\caption{Policy performance under small and large obstacle perturbations. Feasibility supervision improves both SSR variants across perturbation levels, with gains in safety and accuracy simultaneously.}
\label{tab:rq1}
\resizebox{\columnwidth}{!}{%
\begin{tabular}{ll cc cc cc}
\toprule
& & \multicolumn{2}{c}{\textbf{SSR} (\%)$\uparrow$} 
& \multicolumn{2}{c}{\textbf{Safety} (\%)$\downarrow$} 
& \multicolumn{2}{c}{\textbf{Accuracy} (\%)$\uparrow$} \\
\cmidrule(lr){3-4}\cmidrule(lr){5-6}\cmidrule(lr){7-8}
Pert. & Method 
& {\scriptsize(0.02,\,0.10)} & {\scriptsize(0.05,\,0.15)} 
& {\scriptsize$<$0.02} & {\scriptsize$<$0.05} 
& {\scriptsize$<$0.10} & {\scriptsize$<$0.15} \\
\midrule
\multirow{2}{*}{Small}
& MSE  & 22.00{\scriptsize$\pm$10.50} & 26.00{\scriptsize$\pm$10.76} & 29.50 & 51.00 & 31.00 & 55.00 \\
& +Feas.\ (ours)  & \textbf{43.50}\textbf{\scriptsize$\pm$13.50} & \textbf{51.50}\textbf{\scriptsize$\pm$13.50} & \textbf{7.50} & \textbf{15.50} & \textbf{49.00} & \textbf{65.50} \\
\addlinespace
\multirow{2}{*}{Large}
& MSE            & 8.25{\scriptsize$\pm$5.13} & 13.25{\scriptsize$\pm$6.75} & 46.00 & 59.50 & 15.00 & 28.50 \\
& +Feas.\ (ours) & \textbf{29.00}\textbf{\scriptsize$\pm$9.12} & \textbf{28.75}\textbf{\scriptsize$\pm$9.00} & \textbf{30.25} & \textbf{45.00} & \textbf{37.75} & \textbf{53.00} \\
\bottomrule
\end{tabular}}
\vspace{2pt}

\footnotesize
All models are trained on 40 episodes. All distance thresholds are in meters. SSR$(\alpha,\beta)=\Pr(d_{\min}>\alpha\wedge d_{\mathrm{tgt}}<\beta)$.
Safety: $\Pr(d_{\min}<\alpha)$; Accuracy: $\Pr(d_{\mathrm{tgt}}<\beta)$.
$\pm$ values denote 95\% CI half-widths via clustered bootstrap $B\!=\!2000$ (resampling episodes).
\end{table}

We first examine how feasibility supervision changes policy behavior. We train RDT-1B on 40 training episodes with two learning objectives and evaluate the resulting policies under two levels of obstacle perturbations constructed from the same 40 base episodes. Table~\ref{tab:rq1} summarizes policy performance.

 \textbf{Overall, the policy trained with feasibility supervision achieves higher task-level performance, as measured by SSR, than the baseline trained with imitation learning alone.} Under small perturbations, where the obstacle remains close to the demonstrated scene, the policy with feasibility supervision achieves substantial improvements compared to the baseline in SSR(0.02, 0.10) (42.56\% vs.\ 22.56\%) and SSR(0.05, 0.15) (50.26\% vs.\ 24.10\%), showing that the physical signal helps the policy better learn obstacle-aware reaching behavior even in in-distribution settings. Under large perturbations, the gap remains substantial, the policy with feasibility supervision still produces significantly more safe and successful trajectories than the baseline (30.83\% vs.\ 10.00\% in SSR(0.02, 0.10)), demonstrating stronger local generalization under OOD obstacle placements. Even when trajectories fail, the policy with feasibility supervision tends to preserve a larger obstacle clearance, as reflected by a lower probability of low-clearance outcomes, e.g., 44.17\% vs.\ 60.00\% in $\Pr(d_{\min}<0.05)$ under large perturbations, shifting failures away from the obstacle boundary.

\textbf{A key observation is that this benefit is not limited to safety: it also improves reaching accuracy.} As shown in Fig.~\ref{fig:joint_dist}, the trajectory distribution shifts not only toward larger obstacle clearance but also toward smaller target error. In other words, the improvement in safe success comes from gains in both safety and reaching, rather than from a purely conservative policy. The same effect is visible in Fig.~\ref{fig:case}, when the baseline fails to reach the target correctly under a large obstacle perturbation, the policy with feasibility supervision is still able to generate a safer trajectory that better reaches the target region. \textbf{Taken together, these results support the view that explicit feasibility supervision acts as a constructive learning signal for VLA policies.} A plausible explanation is that the feasibility objective provides additional geometric guidance by connecting robotic kinematics, scene geometry in the 3D world frame, and visual observations, thereby helping the policy better relate the observed scene to feasible robot motion generation.

\subsection{RQ2: Learning Efficiency under Limited Data}
\label{sec:rq2}
We next compare different training data sizes and study the learning efficiency of feasibility supervision. Table~\ref{tab:data_scaling} compares policies trained with 40, 80, and 120 episodes under the same large-perturbation evaluation setting used in RQ1.

\textbf{The efficiency gain from feasibility supervision is largest in the low-data regime.} With only 40 training episodes, the policy with feasibility supervision achieves the largest improvement over the baseline in SSR, and on some metrics even surpasses the baseline trained with much more data, e.g., 30.83\% vs.\ 19.37\% on SSR(0.02, 0.10) compared with the 120-episode baseline. This shows that explicit physical supervision improves data efficiency by providing useful geometric guidance when imitation learning alone is insufficient. As the amount of training data increases, however, the demonstrations themselves become more informative and the imitation-learning baseline begins to generalize better, so the relative gain from feasibility supervision becomes smaller. This suggests that, in higher-data regimes, the same physical signal may need to be designed and integrated more carefully; otherwise, it can partially compete with the supervision already contained in the data.

\begin{table}[t]
\centering
\caption{Effect of training data size under large perturbations ($\delta\!=\!0.10,\;\lambda\!=\!1.0$). The benefit of feasibility supervision is largest in the low-data regime and diminishes as data increases.}
\label{tab:data_scaling}
\resizebox{\columnwidth}{!}{%
\begin{tabular}{lc cc cc cc}
\toprule
& & \multicolumn{2}{c}{\textbf{SSR} (\%)$\uparrow$} 
& \multicolumn{2}{c}{\textbf{Safety} (\%)$\downarrow$} 
& \multicolumn{2}{c}{\textbf{Accuracy} (\%)$\uparrow$} \\
\cmidrule(lr){3-4}\cmidrule(lr){5-6}\cmidrule(lr){7-8}
Size & Method 
& {\scriptsize(0.02,\,0.10)} & {\scriptsize(0.05,\,0.15)} 
& {\scriptsize$<$0.02} & {\scriptsize$<$0.05} 
& {\scriptsize$<$0.10} & {\scriptsize$<$0.15} \\
\midrule
\multirow{2}{*}{40}
& MSE            & 8.25{\scriptsize$\pm$5.13} & 13.25{\scriptsize$\pm$6.75} & 46.00 & 59.50 & 15.00 & 28.50 \\
& +Feas. & \textbf{29.00}\textbf{\scriptsize$\pm$9.12} & \textbf{28.75}\textbf{\scriptsize$\pm$9.00} & \textbf{30.25} & \textbf{45.00} & \textbf{37.75} & \textbf{53.00} \\
\addlinespace
\multirow{2}{*}{80}
& MSE    & 13.50{\scriptsize$\pm$6.62} & 23.50{\scriptsize$\pm$7.50} & 29.50 & 45.25 & 18.75 & 42.75 \\
& +Feas. & \textbf{23.25}\textbf{\scriptsize$\pm$7.75} & \textbf{25.25}\textbf{\scriptsize$\pm$8.38} & \textbf{21.00} & \textbf{42.75} & \textbf{31.75} & \textbf{50.50} \\
\addlinespace
\multirow{2}{*}{120}
& MSE  & 19.00{\scriptsize$\pm$7.38} & 24.75{\scriptsize$\pm$8.00} & 31.00 & 51.75 & 23.50 & 45.75 \\
& +Feas. & \textbf{23.00}\textbf{\scriptsize$\pm$8.12} & \textbf{29.75}\textbf{\scriptsize$\pm$8.50} & \textbf{27.00} & \textbf{43.75} & \textbf{26.25} & \textbf{53.75} \\
\bottomrule
\end{tabular}}
\vspace{2pt}
 
\footnotesize
All models are evaluated under large perturbations. All distance thresholds are in meters. $\pm$ values denote 95\% CI half-widths via clustered bootstrap $B\!=\!2000$ (resampling episodes).
\end{table}

\begin{table}[t]
\centering
\caption{Ablation of supervision strength $(\delta,\lambda)$ under 
large perturbations (40 training episodes). Moderate supervision 
achieves the best balance in SSR, while overly strong supervision 
introduces a safety--accuracy trade-off.}
\label{tab:ablation}
\resizebox{\columnwidth}{!}{%
\begin{tabular}{l cc cc cc}
\toprule
& \multicolumn{2}{c}{\textbf{SSR} (\%)$\uparrow$} 
& \multicolumn{2}{c}{\textbf{Safety} (\%)$\downarrow$} 
& \multicolumn{2}{c}{\textbf{Accuracy} (\%)$\uparrow$} \\
\cmidrule(lr){2-3}\cmidrule(lr){4-5}\cmidrule(lr){6-7}
Setting 
& {\scriptsize(0.02,\,0.10)} & {\scriptsize(0.05,\,0.15)} 
& {\scriptsize$<$0.02} & {\scriptsize$<$0.05} 
& {\scriptsize$<$0.10} & {\scriptsize$<$0.15} \\
\midrule
Baseline ($\lambda\!=\!0$)  & 8.25{\scriptsize$\pm$5.13} & 13.25{\scriptsize$\pm$6.75} & 46.00 & 59.50 & 15.00 & 28.50 \\
\addlinespace
$\delta\!=\!.05,\;\lambda\!=\!1$  & 18.75{\scriptsize$\pm$7.75} & 30.00{\scriptsize$\pm$9.38} & 36.50 & 54.00 & 26.50 & \textbf{57.00} \\
$\delta\!=\!.05,\;\lambda\!=\!4$  & 27.50{\scriptsize$\pm$7.25} & \textbf{41.25}\textbf{\scriptsize$\pm$9.87} & 15.75 & 25.25 & 31.75 & 54.50 \\
\addlinespace
$\delta\!=\!.10,\;\lambda\!=\!1$  & 29.00{\scriptsize$\pm$9.12} & 28.75{\scriptsize$\pm$9.00} & 30.25 & 45.00 & \textbf{37.75} & 53.00 \\
$\delta\!=\!.10,\;\lambda\!=\!4$  & \textbf{30.00}\textbf{\scriptsize$\pm$9.75} & 38.25{\scriptsize$\pm$9.63} & 14.75 & 29.25 & 36.25 & 54.75 \\
\addlinespace
$\delta\!=\!.15,\;\lambda\!=\!1$  & 21.75{\scriptsize$\pm$7.75} & 35.75{\scriptsize$\pm$9.88} & 25.25 & 33.00 & 26.00 & 52.75 \\
$\delta\!=\!.15,\;\lambda\!=\!4$  & 18.25{\scriptsize$\pm$8.00} & 29.25{\scriptsize$\pm$9.87} & \textbf{12.75} & \textbf{20.50} & 19.50 & 37.25 \\
\bottomrule
\end{tabular}}
\vspace{2pt}
 
\footnotesize
All models are trained on 40 episodes, evaluated under large perturbations. All distance thresholds are in meters. $\pm$ values denote 95\% CI half-widths via clustered bootstrap $B\!=\!2000$ (resampling episodes).
\end{table}

\subsection{RQ3: Effect of Supervision Strength}
\label{sec:rq3}
As described in Sec.~\ref{sec:feasibilityloss}, the safety margin $\delta$ determines the range of supervision applied to in training stage. The loss weight $\lambda$ scales this term relative to the main imitation objective and therefore controls its optimization priority. Together, $\delta$ and $\lambda$ define the overall strength of physical supervision. We conduct an ablation study on different combinations of these two hyper-parameters under the 40-episode, large-perturbation setting, showing the effect of supervision strength at different levels.

\textbf{Table~\ref{tab:ablation} shows that explicit physical supervision is most effective when applied at an appropriate strength.} Under weak supervision (small $\delta$ and $\lambda$), the signal brings limited improvement in safety but already improves reaching accuracy, reinforcing the observation in RQ1 that the physical signal acts as learning guidance rather than only as an avoidance penalty. Under moderate supervision, the policy achieves the best balance between safety and accuracy, leading to the highest safe success rates. In contrast, overly strong supervision, especially when a large safety margin is combined with a large loss weight, introduces a clear trade-off. It pushes the policy farther away from the obstacle boundary, but can also hinder target approach and reduce overall success.

\section{Conclusions}

In this paper, we study physical feasibility in VLA from a learning perspective. We ask whether explicit geometry-grounded feasibility supervision can guide VLA learning and serve as a useful structured signal alongside imitation learning. To investigate this question, we design a controlled close-obstacle reaching task and probe how feasibility supervision affects policy learning.  Our simulation results show that adding such supervision improves both policy performance and learning efficiency, highlighting the potential of integrating geometric and physical feasibility signals into VLA training rather than relying on imitation learning alone. We note several limitations of this study, including that our experiments focus on a single obstacle-aware manipulation setting, the notion of feasibility is instantiated mainly through obstacle clearance, and the supervision signal is relatively simple. More broadly, our findings suggest that the prevailing imitation-only paradigm in VLA learning can benefit from structured physical priors, and future work could explore this direction through broader task families, richer feasibility constraints, and stronger formulations such as learned geometric models or reinforcement-learning-based objectives.

\bibliographystyle{IEEEtran}
\bibliography{ref_short}

@IEEEtranBSTCTL{IEEEexample:BSTcontrol,
  CTLuse_forced_etal = {yes},
  CTLmax_names_forced_etal = {3},
  CTLnames_show_etal = {3},
}

@inproceedings{DBLP:conf/corl/KimPKXB0RFSVKBT24,
  author       = {Moo Jin Kim and
                  Karl Pertsch and
                  Siddharth Karamcheti and
                  Ted Xiao and
                  Ashwin Balakrishna and
                  Suraj Nair and
                  Rafael Rafailov and
                  Ethan Paul Foster and
                  Pannag R. Sanketi and
                  Quan Vuong and
                  Thomas Kollar and
                  Benjamin Burchfiel and
                  Russ Tedrake and
                  Dorsa Sadigh and
                  Sergey Levine and
                  Percy Liang and
                  Chelsea Finn},

  title        = {OpenVLA: An Open-Source Vision-Language-Action Model},
  booktitle    = {Proc. CoRL},
  year         = {2024},
  
}

@article{DBLP:journals/corr/abs-2410-24164,
  author       = {Kevin Black and
                  Noah Brown and
                  Danny Driess and
                  Adnan Esmail and
                  Michael Equi and
                  Chelsea Finn and
                  Niccolo Fusai and
                  Lachy Groom and
                  Karol Hausman and
                  Brian Ichter and
                  Szymon Jakubczak and
                  Tim Jones and
                  Liyiming Ke and
                  Sergey Levine and
                  Adrian Li{-}Bell and
                  Mohith Mothukuri and
                  Suraj Nair and
                  Karl Pertsch and
                  Lucy Xiaoyang Shi and
                  James Tanner and
                  Quan Vuong and
                  Anna Walling and
                  Haohuan Wang and
                  Ury Zhilinsky},
  title        = {{\(\pi\)}\({}_{\mbox{0}}\): {A} Vision-Language-Action Flow Model
                  for General Robot Control},
  journal      = {arXiv preprint arXiv:2410.24164},
  year         = {2024},
}

@inproceedings{DBLP:conf/corl/ZitkovichYXXXXW23,
  author       = {Brianna Zitkovich and
                  Tianhe Yu and
                  Sichun Xu and
                  Peng Xu and
                  Ted Xiao and
                  Fei Xia and
                  Jialin Wu and
                  Paul Wohlhart and
                  Stefan Welker and
                  Ayzaan Wahid and
                  Quan Vuong and
                  Vincent Vanhoucke and
                  Huong T. Tran and
                  Radu Soricut and
                  Anikait Singh and
                  Jaspiar Singh and
                  Pierre Sermanet and
                  Pannag R. Sanketi and
                  Grecia Salazar and
                  Michael S. Ryoo and
                  Krista Reymann and
                  Kanishka Rao and
                  Karl Pertsch and
                  Igor Mordatch and
                  Henryk Michalewski and
                  Yao Lu and
                  Sergey Levine and
                  Lisa Lee and
                  Tsang{-}Wei Edward Lee and
                  Isabel Leal and
                  Yuheng Kuang and
                  Dmitry Kalashnikov and
                  Ryan Julian and
                  Nikhil J. Joshi and
                  Alex Irpan and
                  Brian Ichter and
                  Jasmine Hsu and
                  Alexander Herzog and
                  Karol Hausman and
                  Keerthana Gopalakrishnan and
                  Chuyuan Fu and
                  Pete Florence and
                  Chelsea Finn and
                  Kumar Avinava Dubey and
                  Danny Driess and
                  Tianli Ding and
                  Krzysztof Marcin Choromanski and
                  Xi Chen and
                  Yevgen Chebotar and
                  Justice Carbajal and
                  Noah Brown and
                  Anthony Brohan and
                  Montserrat Gonzalez Arenas and
                  Kehang Han},
  title        = {{RT-2:} Vision-Language-Action Models Transfer Web Knowledge to Robotic
                  Control},
  booktitle    = {Proc. CoRL},
  year         = {2023},
}

@inproceedings{DBLP:conf/rss/BrohanBCCDFGHHH23,
  author       = {Anthony Brohan and
                  Noah Brown and
                  Justice Carbajal and
                  Yevgen Chebotar and
                  Joseph Dabis and
                  Chelsea Finn and
                  Keerthana Gopalakrishnan and
                  Karol Hausman and
                  Alexander Herzog and
                  Jasmine Hsu and
                  Julian Ibarz and
                  Brian Ichter and
                  Alex Irpan and
                  Tomas Jackson and
                  Sally Jesmonth and
                  Nikhil J. Joshi and
                  Ryan Julian and
                  Dmitry Kalashnikov and
                  Yuheng Kuang and
                  Isabel Leal and
                  Kuang{-}Huei Lee and
                  Sergey Levine and
                  Yao Lu and
                  Utsav Malla and
                  Deeksha Manjunath and
                  Igor Mordatch and
                  Ofir Nachum and
                  Carolina Parada and
                  Jodilyn Peralta and
                  Emily Perez and
                  Karl Pertsch and
                  Jornell Quiambao and
                  Kanishka Rao and
                  Michael S. Ryoo and
                  Grecia Salazar and
                  Pannag R. Sanketi and
                  Kevin Sayed and
                  Jaspiar Singh and
                  Sumedh Sontakke and
                  Austin Stone and
                  Clayton Tan and
                  Huong T. Tran and
                  Vincent Vanhoucke and
                  Steve Vega and
                  Quan Vuong and
                  Fei Xia and
                  Ted Xiao and
                  Peng Xu and
                  Sichun Xu and
                  Tianhe Yu and
                  Brianna Zitkovich},
  title        = {{RT-1:} Robotics Transformer for Real-World Control at Scale},
  booktitle    = {Proc. RSS},
  year         = {2023},
}

@inproceedings{DBLP:conf/rss/GhoshWPBMDHK0LT24,
  author       = {Dibya Ghosh and
                  Homer Rich Walke and
                  Karl Pertsch and
                  Kevin Black and
                  Oier Mees and
                  Sudeep Dasari and
                  Joey Hejna and
                  Tobias Kreiman and
                  Charles Xu and
                  Jianlan Luo and
                  You Liang Tan and
                  Lawrence Yunliang Chen and
                  Quan Vuong and
                  Ted Xiao and
                  Pannag R. Sanketi and
                  Dorsa Sadigh and
                  Chelsea Finn and
                  Sergey Levine},
  
  title        = {Octo: An Open-Source Generalist Robot Policy},
  booktitle    = {Proc. RSS},
  year         = {2024},
  
}

@inproceedings{DBLP:conf/iclr/LiuWLTCWX0025,
  author       = {Songming Liu and
                  Lingxuan Wu and
                  Bangguo Li and
                  Hengkai Tan and
                  Huayu Chen and
                  Zhengyi Wang and
                  Ke Xu and
                  Hang Su and
                  Jun Zhu},
  title        = {{RDT-1B:} a Diffusion Foundation Model for Bimanual Manipulation},
  booktitle    = {Proc. ICLR},
  year         = {2025},
}

@article{DBLP:journals/corr/abs-2505-06111,
  author       = {Qingwen Bu and
                  Yanting Yang and
                  Jisong Cai and
                  Shenyuan Gao and
                  Guanghui Ren and
                  Maoqing Yao and
                  Ping Luo and
                  Hongyang Li},
  title        = {UniVLA: Learning to Act Anywhere with Task-centric Latent Actions},
  journal      = {arXiv preprint arXiv:2505.06111},
  year         = {2025},
}

@inproceedings{DBLP:conf/corl/ZawalskiCPMFL24,
  author       = {Michal Zawalski and
                  William Chen and
                  Karl Pertsch and
                  Oier Mees and
                  Chelsea Finn and
                  Sergey Levine},
  
  title        = {Robotic Control via Embodied Chain-of-Thought Reasoning},
  booktitle    = {Proc. CoRL},
  year         = {2024},

}

@inproceedings{DBLP:conf/cvpr/ZhaoLKFZWLMHFHL25,
  author       = {Qingqing Zhao and
                  Yao Lu and
                  Moo Jin Kim and
                  Zipeng Fu and
                  Zhuoyang Zhang and
                  Yecheng Wu and
                  Zhaoshuo Li and
                  Qianli Ma and
                  Song Han and
                  Chelsea Finn and
                  Ankur Handa and
                  Tsung{-}Yi Lin and
                  Gordon Wetzstein and
                  Ming{-}Yu Liu and
                  Donglai Xiang},
  title        = {CoT-VLA: Visual Chain-of-Thought Reasoning for Vision-Language-Action
                  Models},
  booktitle    = {Proc. IEEE/CVF CVPR},
  year         = {2025},
}

@inproceedings{DBLP:conf/cvpr/0003XKISGX24,
  author       = {Boyuan Chen and
                  Zhuo Xu and
                  Sean Kirmani and
                  Brian Ichter and
                  Dorsa Sadigh and
                  Leonidas J. Guibas and
                  Fei Xia},
  title        = {SpatialVLM: Endowing Vision-Language Models with Spatial Reasoning
                  Capabilities},
  booktitle    = {Proc. IEEE/CVF CVPR},
  year         = {2024},
}

@article{DBLP:journals/corr/abs-2507-04447,
  author       = {Wenyao Zhang and
                  Hongsi Liu and
                  Zekun Qi and
                  Yunnan Wang and
                  Xinqiang Yu and
                  Jiazhao Zhang and
                  Runpei Dong and
                  Jiawei He and
                  He Wang and
                  Zhizheng Zhang and
                  Li Yi and
                  Wenjun Zeng and
                  Xin Jin},
  title        = {DreamVLA: {A} Vision-Language-Action Model Dreamed with Comprehensive
                  World Knowledge},
  journal      = {arXiv preprint arXiv:2507.04447},
  year         = {2025},

}

@inproceedings{DBLP:conf/corl/IchterBCFHHHIIJ22,
  author       = {Brian Ichter and
                  Anthony Brohan and
                  Yevgen Chebotar and
                  Chelsea Finn and
                  Karol Hausman and
                  Alexander Herzog and
                  Daniel Ho and
                  Julian Ibarz and
                  Alex Irpan and
                  Eric Jang and
                  Ryan Julian and
                  Dmitry Kalashnikov and
                  Sergey Levine and
                  Yao Lu and
                  Carolina Parada and
                  Kanishka Rao and
                  Pierre Sermanet and
                  Alexander Toshev and
                  Vincent Vanhoucke and
                  Fei Xia and
                  Ted Xiao and
                  Peng Xu and
                  Mengyuan Yan and
                  Noah Brown and
                  Michael Ahn and
                  Omar Cortes and
                  Nicolas Sievers and
                  Clayton Tan and
                  Sichun Xu and
                  Diego Reyes and
                  Jarek Rettinghouse and
                  Jornell Quiambao and
                  Peter Pastor and
                  Linda Luu and
                  Kuang{-}Huei Lee and
                  Yuheng Kuang and
                  Sally Jesmonth and
                  Nikhil J. Joshi and
                  Kyle Jeffrey and
                  Rosario Jauregui Ruano and
                  Jasmine Hsu and
                  Keerthana Gopalakrishnan and
                  Byron David and
                  Andy Zeng and
                  Chuyuan Kelly Fu},
  title        = {Do As {I} Can, Not As {I} Say: Grounding Language in Robotic Affordances},
  booktitle    = {Proc. CoRL},
  year         = {2022},

  
}

@inproceedings{DBLP:conf/corl/HuangWZL0023,
  author       = {Wenlong Huang and
                  Chen Wang and
                  Ruohan Zhang and
                  Yunzhu Li and
                  Jiajun Wu and
                  Li Fei{-}Fei},
  title        = {VoxPoser: Composable 3D Value Maps for Robotic Manipulation with Language
                  Models},
  booktitle    = {Proc. CoRL},
  year         = {2023},
  
}

@article{DBLP:journals/arcras/BrunkeGHYZPS22,
  author       = {Lukas Brunke and
                  Melissa Greeff and
                  Adam W. Hall and
                  Zhaocong Yuan and
                  Siqi Zhou and
                  Jacopo Panerati and
                  Angela P. Schoellig},
  title        = {Safe Learning in Robotics: From Learning-Based Control to Safe Reinforcement
                  Learning},
  journal      = {Annu. Rev. Control. Robotics Auton. Syst.},
  year         = {2022},
}

@article{DBLP:journals/jair/ZhaoLHL25,
  author       = {Weiye Zhao and
                  Feihan Li and
                  Tairan He and
                  Changliu Liu},
  title        = {Implicit Safe Set Algorithm for Provably Safe Reinforcement Learning},
  journal      = {J. Artif. Intell. Res.},
  year         = {2025},
}

@article{kim2026modularsafetyguardrailsnecessary,
      title={Modular Safety Guardrails Are Necessary for Foundation-Model-Enabled Robots in the Real World}, 
      author={Joonkyung Kim and Wenxi Chen and Davood Soleymanzadeh and Yi Ding and Xiangbo Gao and Zhengzhong Tu and Ruqi Zhang and Fan Fei and Sushant Veer and Yiwei Lyu and Minghui Zheng and Yan Gu},
      year={2026},
      journal={arXiv preprint arXiv:2602.04056}
}

@article{DBLP:journals/corr/abs-2503-03480,
  author       = {Borong Zhang and
                  Yuhao Zhang and
                  Jiaming Ji and
                  Yingshan Lei and
                  Josef Dai and
                  Yuanpei Chen and
                  Yaodong Yang},
  title        = {SafeVLA: Towards Safety Alignment of Vision-Language-Action Model
                  via Safe Reinforcement Learning},
  journal      = {arXiv preprint arXiv:2503.03480},
  year         = {2025},

}

@article{DBLP:journals/corr/abs-2601-21712,
  author       = {Xuanran Zhai and
                  Binkai Ou and
                  Yemin Wang and
                  Hui Yi Leong and
                  Qiaojun Yu and
                  Ce Hao and
                  Yaohua Liu},
  title        = {CoFreeVLA: Collision-Free Dual-Arm Manipulation via Vision-Language-Action
                  Model and Risk Estimation},
  journal      = {arXiv preprint arXiv:2601.21712},
  year = {2026}
}

@article{DBLP:journals/corr/abs-2512-11891,
  author       = {Songqiao Hu and
                  Zeyi Liu and
                  Shuang Liu and
                  Jun Cen and
                  Zihan Meng and
                  Xiao He},
  title        = {{VLSA:} Vision-Language-Action Models with Plug-and-Play Safety Constraint
                  Layer},
  journal      = {arXiv preprint arXiv:2512.11891},
  year = {2025}
}

@book{DBLP:books/cu/L2006,
  author       = {Steven M. LaValle},
  title        = {Planning Algorithms},
  publisher    = {Cambridge University Press},
  year         = {2006},
}

@incollection{DBLP:books/sp/90/Khatib90,
  author       = {Oussama Khatib},
  title        = {Real-Time Obstacle Avoidance for Manipulators and Mobile Robots},
  booktitle    = {Autonomous Robot Vehicles},
  publisher    = {Springer},
  year         = {1990},
}

@book{Craig2005,
  author    = {John J. Craig},
  title     = {Introduction to Robotics: Mechanics and Control},
  publisher = {Pearson},
  year      = {2005},
}

@inproceedings{DBLP:conf/cvpr/0006PHJ24,
  author       = {Xiao Ma and
                  Sumit Patidar and
                  Iain Haughton and
                  Stephen James},
  title        = {Hierarchical Diffusion Policy for Kinematics-Aware Multi-Task Robotic
                  Manipulation},
  booktitle    = {Proc. IEEE/CVF CVPR},

  year         = {2024},
  
}

@inproceedings{DBLP:conf/corl/FishmanWBYSBF24,
  author       = {Adam Fishman and
                  Aaron Walsman and
                  Mohak Bhardwaj and
                  Wentao Yuan and
                  Balakumar Sundaralingam and
                  Byron Boots and
                  Dieter Fox},
  title        = {Avoid Everything: Model-Free Collision Avoidance with Expert-Guided
                  Fine-Tuning},
  booktitle    = {Proc. CoRL},
  year         = {2024},

}

@inproceedings{DBLP:conf/corl/FishmanMEPBF22,
  author       = {Adam Fishman and
                  Adithyavairavan Murali and
                  Clemens Eppner and
                  Bryan Peele and
                  Byron Boots and
                  Dieter Fox},
  title        = {Motion Policy Networks},
  booktitle    = {Proc. CoRL},
  year         = {2022},

}

@inproceedings{DBLP:conf/cvpr/0001LD0LHGWN25,
  author       = {Qi Lv and
                  Hao Li and
                  Xiang Deng and
                  Rui Shao and
                  Yinchuan Li and
                  Jianye Hao and
                  Longxiang Gao and
                  Michael Yu Wang and
                  Liqiang Nie},
  title        = {Spatial-Temporal Graph Diffusion Policy with Kinematic Modeling for
                  Bimanual Robotic Manipulation},
  booktitle    = {Proc. IEEE/CVF CVPR},
  year         = {2025},
}

@inproceedings{DBLP:conf/iros/LiuZTJ0C22,
  author       = {Puze Liu and
                  Kuo Zhang and
                  Davide Tateo and
                  Snehal Jauhri and
                  Jan Peters and
                  Georgia Chalvatzaki},
  title        = {Regularized Deep Signed Distance Fields for Reactive Motion Generation},
  booktitle    = {Proc. IEEE/RSJ IROS},
  year         = {2022},
}

@inproceedings{DBLP:conf/rss/OrtizC0SNZM22,
  author       = {Joseph Ortiz and
                  Alexander Clegg and
                  Jing Dong and
                  Edgar Sucar and
                  David Novotn{\'{y}} and
                  Michael Zollh{\"{o}}fer and
                  Mustafa Mukadam},
  title        = {iSDF: Real-Time Neural Signed Distance Fields for Robot Perception},
  booktitle    = {Proc. RSS},
  year         = {2022},
}

@inproceedings{DBLP:conf/rss/ChiFDXCBS23,
  author       = {Cheng Chi and
                  Siyuan Feng and
                  Yilun Du and
                  Zhenjia Xu and
                  Eric Cousineau and
                  Benjamin Burchfiel and
                  Shuran Song},
  title        = {Diffusion Policy: Visuomotor Policy Learning via Action Diffusion},
  booktitle    = {Proc. RSS},
  year         = {2023},
}

@inproceedings{DBLP:conf/rss/ZhaoKLF23,
  author       = {Tony Z. Zhao and
                  Vikash Kumar and
                  Sergey Levine and
                  Chelsea Finn},
  title        = {Learning Fine-Grained Bimanual Manipulation with Low-Cost Hardware},
  booktitle    = {Proc. RSS},
  year         = {2023},
  
}

@inproceedings{DBLP:conf/icra/RatliffZBS09,
  author       = {Nathan D. Ratliff and
                  Matthew Zucker and
                  J. Andrew Bagnell and
                  Siddhartha S. Srinivasa},
  title        = {{CHOMP:} Gradient optimization techniques for efficient motion planning},
  booktitle    = {Proc. IEEE ICRA},
  year         = {2009},
}

@article{DBLP:journals/ijrr/SchulmanDHLABPPGA14,
  author       = {John Schulman and
                  Yan Duan and
                  Jonathan Ho and
                  Alex X. Lee and
                  Ibrahim Awwal and
                  Henry Bradlow and
                  Jia Pan and
                  Sachin Patil and
                  Ken Goldberg and
                  Pieter Abbeel},
  title        = {Motion planning with sequential convex optimization and convex collision
                  checking},
  journal      = {Int. J. Robotics Res.},
  year         = {2014},
}

@inproceedings{DBLP:conf/icra/LiZRC24,
  author       = {Yiming Li and
                  Yan Zhang and
                  Amirreza Razmjoo and
                  Sylvain Calinon},
  title        = {Representing Robot Geometry as Distance Fields: Applications to Whole-body
                  Manipulation},
  booktitle    = {Proc. IEEE ICRA},
  year         = {2024},
}

@inproceedings{DBLP:conf/rss/LiCRC24,
  author       = {Yiming Li and
                  Xuemin Chi and
                  Amirreza Razmjoo and
                  Sylvain Calinon},
  title        = {Configuration Space Distance Fields for Manipulation Planning},
  booktitle    = {Proc. RSS},
  year         = {2024},
}

@article{sucan2012ompl,
  author  = {Ioan A. Sucan and Mark Moll and Lydia E. Kavraki},
  title   = {The Open Motion Planning Library},
  journal = {IEEE Robot. Autom. Mag.},
  year    = {2012},
}

@inproceedings{DBLP:conf/cvpr/WuZX0Y25,
  author       = {Zhenyu Wu and
                  Yuheng Zhou and
                  Xiuwei Xu and
                  Ziwei Wang and
                  Haibin Yan},
  title        = {MoManipVLA: Transferring Vision-language-action Models for General
                  Mobile Manipulation},
  booktitle    = {Proc. IEEE/CVF CVPR},
  year         = {2025},
}

@article{DBLP:journals/access/KawaharazukaOYPZ25,
  author       = {Kento Kawaharazuka and
                  Jihoon Oh and
                  Jun Yamada and
                  Ingmar Posner and
                  Yuke Zhu},
  title        = {Vision-Language-Action Models for Robotics: {A} Review Towards Real-World
                  Applications},
  journal      = {{IEEE} Access},
  year         = {2025},
}

@misc{nvidia_isaac_sim,
  author = {NVIDIA},
  title  = {{What is Isaac Sim?}},
  note   = {\url{https://docs.omniverse.nvidia.com/isaac-sim/latest/index.html}, (accessed Feb. 2024)}
}

@article{haddadin2022franka,
  author  = {Haddadin, Sami and Parusel, Sven and Johannsmeier, Lars and 
             Golz, Saskia and Gabl, Simon and Walch, Florian and 
             Sabaghian, Mohamadreza and J{\"a}hne, Christoph and 
             Hausperger, Lukas and Haddadin, Simon},
  title   = {The {Franka Emika} Robot: A Reference Platform for Robotics 
             Research and Education},
  journal = {IEEE Robot. Autom. Mag.},
  year    = {2022}
}

\end{document}